\documentclass[journal]{IEEEtran}  % Comment this line out if you need a4paper

\usepackage[dvipsnames]{xcolor}
\usepackage{microtype}
\usepackage{url}
\usepackage{graphicx}
\usepackage{amsmath}
\usepackage{booktabs}
\usepackage{makecell}
\usepackage{multirow}
\usepackage{tikz}
\usetikzlibrary{shapes,positioning}
\usetikzlibrary{mindmap}
\usetikzlibrary{backgrounds, fit, decorations.pathreplacing,}
\usetikzlibrary{matrix}
\usepackage{amsmath,amsfonts}
\usepackage{algorithmic}
\usepackage{algorithm}
\usepackage{array}
\usepackage{subcaption}
\usepackage{textcomp}
\usepackage{url}
\usepackage{hyperref}
\usepackage[capitalize]{cleveref}
\usepackage{verbatim}
\usepackage{graphicx}
\usepackage{orcidlink}
\usepackage[hyperref=true, backref=false, sorting=none]{biblatex}
\usepackage{colortbl}
\colorlet{colorFst}{Green!40}
\colorlet{colorSnd}{SpringGreen!50} % second

\definecolor{backgroundColor}{HTML}{b0aea4}
\definecolor{modelColor}{HTML}{345995}

\tikzset{
 base node/.style={
 rounded corners=3pt,
 align=center,
 minimum height=0.5cm,
 inner sep=8pt
 },
 category box/.style={
 rectangle,
 rounded corners,
 draw=gray!30,
 fill=gray!5,
 inner sep=2.5pt
 },
 category output/.style={
    rectangle,
    rounded corners,
    dashed,
    draw=gray!30,
    fill=gray!5,
    inner sep=2.5pt
    },
 arrow/.style={
 -latex,
 ultra thick,
 opacity=0.9
 },
 models/.style={base node, fill=modelColor!5, draw=modelColor},
 samples1/.style={base node, fill=red!5, draw=red},
 samplesn/.style={base node, fill=teal!5, draw=teal},
 basestyle/.style={base node, fill=backgroundColor!5, draw},
}

\pgfdeclarelayer{background}
\pgfdeclarelayer{foreground}
\pgfsetlayers{background,main,foreground}

\begin{document}

% \title{UAC: \underline{u}ncertainty-\underline{a}ware \underline{c}alibration of neural networks for gesture detection from IMU data in out-of-distribution scenarios}

\title{UAC: \underline{u}ncertainty-\underline{a}ware \underline{c}alibration of neural networks for gesture detection}

\author{Farida Al Haddad\orcidlink{0009-0004-2566-4804}, Yuxin Wang\orcidlink{0009-0000-1187-1037}, and Malcolm Mielle\orcidlink{0000-0002-3079-0512}% <-this % stops a space
    %     \thanks{*This work was supported by Schindler AG}% <-this % stops a space
    \thanks{Farida Al Haddad and Yuxin Wang are with Ecole Polytechnique Federale de Lausanne, Lausanne, Switzerland.
            {\tt\small
                al.farida@epfl.ch, wangyuxin\_99@hotmail.com    }}%
    \thanks{Malcolm Mielle is with Schindler EPFL Lab, Lausanne, Switzerland
            {\tt\small malcolm.mielle@ik.me}}%
}

% \markboth{IEEE Transactions on pattern analysis and machine intelligence}%


\maketitle

\begin{abstract}
    Artificial intelligence has the potential to impact safety and efficiency in safety-critical domains such as construction, manufacturing, and healthcare.
    For example, using sensor data from wearable devices, such as inertial measurement units (IMUs), human gestures can be detected while maintaining privacy, thereby ensuring that safety protocols are followed.
    However, strict safety requirements in these domains have limited the adoption of AI, since accurate calibration of predicted probabilities and robustness against out-of-distribution (OOD) data is necessary.
    This paper proposes UAC (Uncertainty-Aware Calibration), a novel two-step method to address these challenges in IMU-based gesture recognition.
    First, we present an uncertainty-aware gesture network architecture that predicts both gesture probabilities and their associated uncertainties from IMU data.
    This uncertainty is then used to calibrate the probabilities of each potential gesture.
    Second, an entropy-weighted expectation of predictions over multiple IMU data windows is used to improve accuracy while maintaining correct calibration.
    Our method is evaluated using three publicly available IMU datasets for gesture detection and is compared to three state-of-the-art calibration methods for neural networks: temperature scaling, entropy maximization, and Laplace approximation.
    UAC outperforms existing methods, achieving improved accuracy and calibration in both OOD and in-distribution scenarios.
    Moreover, we find that, unlike our method, none of the state-of-the-art methods significantly improve the calibration of IMU-based gesture recognition models.
    In conclusion, our work highlights the advantages of uncertainty-aware calibration of neural networks, demonstrating improvements in both calibration and accuracy for gesture detection using IMU data.
    % thanks to uncertainty-aware calibration of neural networks, our work shows significant improvement in both calibration and accuracy on gesture detection using IMU data only.
\end{abstract}


%%%%%%%%%%%%%%%%%%%%%%%%%%%%%%%%%%%%%%%%%%%%%%%%%%%%%%%%%%%%%%%%%%%%%%%%%%%%%%%%
\section{Introduction}

Over the past decade, advancements in gesture recognition technology have significantly transformed various fields, such as human-computer interaction~\cite{9463455}, communication~\cite{8691602}, and healthcare~\cite{su13052961}.
These innovations have been driven by improvements in sensor technologies, machine learning algorithms, and computational power, resulting in improvements in the speed and accuracy of gesture recognition algorithms.
Nevertheless, the adoption of such technologies in safety-critical applications---notably the construction industry---remains limited due to the need to ensure system safety and reliability.
Despite construction ranking as the most hazardous occupational sector within the European Union, with 22.5\% of all work-related accidents,\footnote{\url{https://ec.europa.eu/eurostat/statistics-explained/index.php?title=Accidents_at_work_statistics}} investments in digital and innovative technologies by the construction sector remain low~\cite{european2020renovation}---70\% of construction firms allocate less than 1\% of their revenues to digital and innovative projects.
Gesture recognition technology holds promise in helping to identify hazardous behaviors, for example by identifying non-compliance with safety protocols, health risks (such as. early signs of heat stroke), or inter-worker hazards (i.e. situations where one worker's actions pose risks to others but not to themselves).
Such situational awareness could significantly improve safety and reduce workload and time demands on construction workers.

\begin{figure}[t]
    \centering
    \input{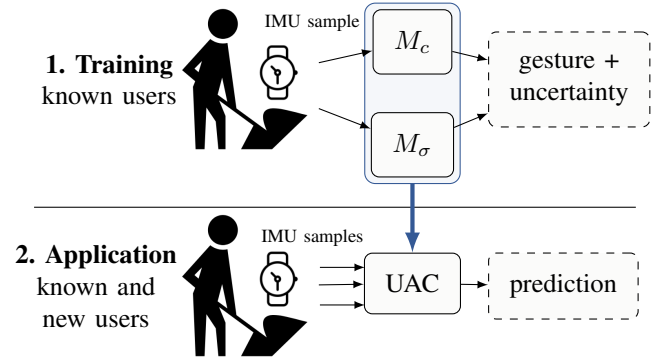}
    \caption{
        Flowchart of the 2-step process of our method.
        1) The model is trained on the training set to detect gestures and estimate the uncertainty associated with the data and the prediction.
        2) The model is used (in out-of or in-distribution scenarios) to aggregate predictions in an uncertainty-aware manner to improve the overall performance while maintaining network calibration.
    }
    \label{fig:intro:image}
\end{figure}

\IEEEpubidadjcol

Typical sensors used for gesture detection include Inertial Measurement Unit (IMU), RGB and RGB-D cameras~\cite{mollyn2022samosa, 10122710}, or even data gloves~\cite{data_glove}.
Those sensor modalities have been used on their own or combined with one another to improve accuracy---for example, \textcite{mollyn2022samosa} have shown the benefits of combining IMU data with audio data, and \textcite{Zou2023} fused vision-based motion signals and sEMG signals using a multi-modal fusion model to achieve high accuracy in hand gesture recognition.
In real-world scenarios however, practical constraints limit sensor choices; while data gloves and surface electromyography (EMG) sensors are costly and inconvenient, cameras may have to be avoided due to privacy concerns, and sensor arrays placed directly in the environments may be rejected due to safety or cost concerns.
On the other hand, IMU sensors integrated into smartwatches are widely accepted by users for their unobtrusiveness and convenience.
Nevertheless, IMU data is less informative than images or video, leading prior research to usually either rely on additional sensor modalities ~\cite{mollyn2022samosa, 10122710} or to focus on in-distribution scenarios~\cite{mollyn2022samosa} to achieve higher accuracy.

The advances in deep learning have led to an increase in the versatility, robustness, and generalization capabilities of gesture recognition.
However, as demonstrated by \textcite{pmlr-v70-guo17a}, deep learning model typically exhibits overconfidence in their predictions and, in safety-critical scenarios, overconfidence or inaccurate predictions can lead to catastrophic, potentially fatal, outcomes.
Thus, accurate probability estimates---i.e. model calibration---are essential for integrating AI into safety-critical scenarios.
Furthermore, the issue of overconfidence and uncertainty of the predictions is exacerbated when predictions are made on individuals not included in the training dataset---a challenge known as Out-Of-Distribution (OOD) domain generalization~\cite{glorot2011domain}.
% Despite extensive research in gesture recognition, existing algorithms still struggle to achieve robust performance.
For the remainder of this paper, OOD will specifically refer to feature shift, where the shift occurs in the input data, as opposed to label shift, where unseen targets may include classes not present in the training data.
In practical scenarios such as construction work, OOD situations are common when a model performs gesture recognition on new workers---an inevitable scenario in real-world applications---leading to degraded model performances and, consequently, increased uncertainty and risk.
Thus, using gesture recognition algorithms based on neural network in safety-critical applications faces three interconnected challenges: the need for high accuracy, calibrated models, and the ability to handle out-of-distribution users.

% \begin{figure}[t]
%     \centering
%     \includegraphics[width=0.48\textwidth]{sections/imgs/IMU_graph.pdf}
%     \caption{
%         Example of OOD detection.
%         The model is trained on a dataset of gestures from a group of workers.
%         When a new worker is introduced, the model is not able to accurately predict the gestures, and the uncertainty is high.
%     }
% \end{figure}

Limited research has been conducted on the calibration of gesture detection models using solely IMU data, particularly in the context of OOD generalization.
In this paper, we focus on two research gaps: 1) accuracy of gesture recognition using a single smartwatch
% 2) uncertainty prediction,
and 2) calibration of the model, in the context of both ID and OOD.
Our method, called \emph{Uncertainty-Aware Calibration} (UAC), is trained and used in two steps.
First, a classification model is trained on short sequences of labeled IMU data to estimate both a gesture prediction and its uncertainty---see the top section of \cref{fig:intro:image}.
The uncertainty is used to perform Monte Carlo integration on the classification model's logits, outputting uncertainty-aware probabilities.
In the second step, the trained classification model is used to predict, on multiple sequences taken from one OOD sample, probabilities of a given gesture for each sequence.
The final sample probability corresponds to the expectation of all sequences' probabilities, weighted by their entropy---see the bottom part of \cref{fig:intro:image}---maintaining calibration of the final prediction while improving the accuracy compared to single sequence prediction.

Thus, the key contributions of this paper are:
\begin{itemize}
    \item We introduce a new method for predicting uncertainty in IMU-based gesture detection, where IMU data is first encoded into a feature space to predict logits and uncertainty.
          Monte Carlo integration is then used to obtain uncertainty-aware probabilities.
    \item We present a novel two-step approach for precise and calibrated gesture detection that incorporates uncertainty at both stages, resulting in enhanced accuracy and calibration compared to the state-of-the-art.
          1) Initially, a network is trained to predict a set of uncertainty-aware probabilities for gesture detection given a short IMU sample.
          2) Multiple predictions from samples extracted from a gesture sequence are combined to derive a final gesture probability.
    \item We conduct a comprehensive evaluation of our method on three publicly available datasets and benchmark our approach against three leading calibration techniques for neural networks: temperature scaling, entropy maximization, and Laplace/Bayesian neural networks.
\end{itemize}

The paper is organized as follows.
\cref{sec:rw} reviews the literature in the domain of gesture recognition and uncertainty in machine learning.
\cref{sec:method} details the method (add details here).
\cref{sec:eval} presents the experimental protocol, the
database, and the classification results.
\cref{sec:discussion} discusses the results and \cref{sec:conclusion} concludes the paper.

Code implementation of the method and evaluation is available online to enable reproducibility of the results.\footnote{\url{https://github.com/Schindler-EPFL-Lab/UAC}}
% \footnote{Link will be added later}

\section{Related Work}
\label{sec:rw}

\subsection{Gesture recognition}
\label{sec:rw:subsec:gesture}

The field of gesture recognition focuses on identifying and interpreting human gestures.
This task can be accomplished using a variety of sensor modalities and data types, including images~\cite{mathe2018arm, jalal2014depth, izadmehr2022depth}, inertial measurement signals~\cite{arduser2016recognizing, kim2019imu, mummadi2018real}, or surface electromyography (EMG) signals~\cite{jiangshuoEMG, joseEMG2021, geng2016gestureEMG}.
Gesture recognition finds applications in diverse domains such as human-robot interaction~\cite{1613091}, sign language recognition~\cite{7916786}, and rehabilitation~\cite{izadmehr2022depth}.

While RGB images have been used for 3D hand-tracking, achieving over 95\% accuracy, methods relying on these sensors are usually sensitive to lighting conditions and occlusion~\cite{zhang2020mediapipe, nohDecadeSurvey}.
Depth-based methods---using devices like the Kinect~\cite{oikonomidis2011efficient}---help improve robustness but require specialized hardware that may not be available.
Electromyography (EMG) signal, which measures muscle activity, has been used for gesture recognition for their non-invasive nature and high accuracy~\cite{jiangshuoEMG, joseEMG2021, geng2016gestureEMG}.
However, EMG-based methods require specialized equipment and are cumbersome for users---for example, requiring that they shave---limiting their practicality.

IMU sensors---consisting of three accelerometers, three gyroscopes, and three magnetometers---are non-invasive and cost-effective sensors, making them suitable for gesture recognition~\cite{mollyn2022samosa, 8323826}
IMU sensors are often integrated into wearable devices, enhancing their applicability across various domains.
Nevertheless, since accelerometer and gyroscopes respectively measure proper acceleration and rate of rotation, IMU data is not as expressive as images or EGM data making gesture detection challenging due to measurement noise, possible similar sensor output among different gestures, and dependency on sensor placement on the body, all of which can lead to misclassification.

To improve the precision of gesture recognition from IMU data, an efficient, albeit simple, strategy is to sum the prediction of a motion over multiple samples and use the prediction with the highest aggregated score as the prediction, as demonstrated by \textcite{mollyn2022samosa}.
However, this method does not provide a set of probabilities and is, therefore, uncalibrated by definition.
% While \textcite{} use delimiter gestures to improve accuracy, such delimiter gestures limit the applicability of the method.
Prior research has also investigated the use of multi-modal sensor inputs to improve the accuracy of gesture recognition from IMU data.
For example, \textcite{mollyn2022samosa} introduce a multi-modal framework that combines IMU and audio data to improve gesture recognition accuracy, and \textcite{DDNN_multimodal} perform simultaneous gesture segmentation and recognition from skeleton data, and RGB and depth images.
However, challenges persist with respect to computational complexity, modality imbalance in fusion, and limited scalability.
% a Gaussian-Bernoulli Deep Belief Network (DBN) for skeletal features, a 3D Convolutional Neural Network (3DCNN) for RGB-D data, and a Hidden Markov Model (HMM) for temporal modeling. The system leverages intermediate and late fusion strategies to integrate modalities, demonstrating complementary strengths—skeleton data excels in segmentation, while RGB-D improves recognition accuracy. Evaluated on the ChaLearn LAP dataset, DDNN achieves a Jaccard index of 0.81, comparable to state-of-the-art methods. However, challenges remain, including computational complexity, modality imbalance in fusion, and limited scalability. The work highlights the potential of deep learning in multimodal gesture recognition and the importance of temporal modeling for continuous streams.
On the other hand, frameworks that combine IMU data with images or videos~\cite{s19091988} have demonstrated more robust results in recognizing various human activities.
While multi-modal approaches generally outperform single-modal approaches, the use of multiple sensors introduces additional technical and material complexity, as well as increased cost.

While IMU sensors are cost-effective and already ubiquitous in our daily lives through smartphones and wearable devices, using them as the sole data source for gesture recognition remains challenging.
There is a need for innovative methods that can effectively leverage IMU data for gesture recognition while addressing the limitations of single-modal approaches, particularly in safety-critical scenarios.

\subsection{Neural Network Calibration}

As seen in \cref{sec:rw:subsec:gesture}, state-of-the-art approaches to gesture recognition predominantly use neural network architectures, such as convolutional neural networks~\cite{mollyn2022samosa}, recurrent neural networks (RNNs)~\cite{8545718}, and long short-term memory (LSTM)~\cite{NEURIPS2018_287e03db}, to improve accuracy.
However, neural networks often exhibit overconfidence in their predictions~\cite{pmlr-v70-guo17a}, meaning that they assign high probabilities to both true and false positive outcomes.
Addressing this overconfidence to better reflect the true uncertainty of the predictions is known as model calibration.

Several strategies for model calibration have been proposed in the literature.
One notable and straightforward method is temperature scaling~\cite{pmlr-v70-guo17a}, where the logits of the neural network are scaled by a temperature parameter $T$.
The temperature parameter is learned post-training during a calibration phase, optimizing $T$ to minimize the cross-entropy loss between the scaled logits and the true labels on a validation set.
Alternatively, \textcite{NEURIPS2021_a7c95857} propose to use Bayesian neural networks (BNNs) for uncertainty estimation and rescaling of the predicted probabilities.
Their approach is either integrated during model training or applied as a fine-tuning step.
\textcite{focal_loss} propose to use the focal loss, which augments the standard cross-entropy loss with a term that emphasizes difficult samples, thereby mitigating the effect of class imbalance.
Focal loss has been shown to reduce overconfidence in model predictions, leading to improved calibration.
\textcite{max-entropy} present a method that maximizes the entropy of incorrect predictions, ensuring that correct predictions have low entropy while incorrect predictions have high entropy.
A key advantage of their approach is that it does not require a separate calibration set.

While the discussed methods have primarily been applied and validated on image-based tasks and datasets, IMU sensors present unique challenges due to their less expressive nature.
As will be demonstrated in \cref{sec:eval}, these existing approaches fall short of effectively calibrating neural networks for gesture detection on IMU sensor data.
Therefore, developing specialized calibration techniques that can effectively harness the unique characteristics of IMU sensor data is crucial for advancing gesture recognition.

\subsection{Out-of-Distribution Generalization}

Beyond the issue of over-confidence, another critical challenge in gesture recognition is the ability to generalize to unseen data, such as new users, new gestures, or new environments.
This capability is particularly important in safety-critical scenarios, where misclassifications can have severe consequences.
Gesture recognition, especially from IMU data, involves time-series data that can vary significantly among subjects due to morphological, physiological, and behavioral factors.
This variability can result in OOD shifts---where the shift occurs in the input data---between training and testing data, negatively impacting model performance and OOD generalization~\cite{general_fw}---i.e., there is a need to ensure that models perform well on unseen data with different distributions.

As shown by \textcite{general_fw}, time-series data are inherently non-stationary, meaning their distribution change over time, leading to distribution shifts.
There are two distinct types of shifts in time-series data: temporal shift and spatial shift.
Temporal shift refers to distribution changes within the same class over time, such as a person walking differently at different times of the day, highlighting the non-stationary nature of time series.
Spatial shift, on the other hand, occurs when the same class exhibits different distributions across sub-populations, such as different users or devices capturing the same activity.

The relationship between OOD performance and model calibration has been shown by \textcite{wald2021calibration}.

\subsection{Uncertainty}

Understanding what a model knows and does not know is crucial for ensuring safe and reliable decision-making in machine learning systems.
While calibration focuses on the accuracy of a model's probability estimations, uncertainty estimation aims to quantify the uncertainty associated with the model's predictions without changing it.
The quantification of uncertainty in machine learning can be categorized into regression and classification tasks.

There are two types of uncertainty that can be modeled:
\begin{enumerate}
    \item \emph{Aleatoric uncertainty}: This refers to the uncertainty inherent to the data, arising from noise and randomness intrinsic to the sensor used to collect data.
    \item \emph{Epistemic uncertainty}: This pertains to the uncertainty associated with the model and its parameters.
          For example, epistemic uncertainty can stem from insufficient information in training data to adequately learn the data distribution.
\end{enumerate}
While epistemic uncertainty can be mitigated with sufficient data, aleatoric uncertainty is irreducible in this manner.
A widely used method to capture model uncertainty is the application of Bayesian Neural Networks (BNNs)~\cite{kendall, NEURIPS2021_a7c95857}, which estimate the posterior distribution over the weights of the neural network.
\textcite{kendall} introduce a Bayesian deep learning framework that explicitly models both aleatoric (data-dependent noise) and epistemic (model uncertainty) uncertainties for vision tasks.

Thus, when looking at neural networks in the context of gesture detection, there is a need to address the uncertainty of the prediction to ensure confidence in the model.
Confidence in the model's prediction is essential for building trustworthy systems, especially in applications where accuracy and safety are paramount.

In conclusion, the existing literature reveals a gap in research on calibration and uncertainty quantification in gesture recognition, particularly within the context of safety-critical applications and OOD scenarios.
To address this, our paper introduces a novel method for tackling calibration and uncertainty quantification in gesture detection.
We evaluate our approach using IMU data and OOD scenarios, where the model encounters new users in the test set who were not part of the training data.

\section{Proposed Method}
\label{sec:method}

\begin{figure*}[t]
    \centering
    \begin{subfigure}[t]{\textwidth}
        \centering
        \resizebox{\textwidth}{!}{\begin{tikzpicture}

    \node[basestyle, text width=0.3cm, align=center] (sample1) at (-3, 4) {$x_i$};

    \begin{scope}[local bounding box=models]

        \node[basestyle, align=center] (encoder) at (-1.25, 4) {Encoder};
        \node[samples1, text width=2cm, align=center] (model) at (1.5, 4.95) {Classification model ($M_c$)};
        \node[samples1, text width=1.5cm, align=center, right= 0.5cm of model, dashed] (logit) {logits $f_W(x_i)$};

        \node[samplesn, text width=2cm, align=center, below= 0.6cm of model] (model_u) { Uncertainty model ($M_\sigma$)};
        \node[samplesn, text width=1.5cm, align=center, right= 0.5cm of model_u, dashed] (sigma) {uncertainty $\sigma(x_i)^2$};
        \node[basestyle, text width=1.85cm, align=center] (markov) at (7, 4) {Monte Carlo Integration};
        \node[text=modelColor] at (7.6, 5.6) {$M_u$};
    \end{scope}

    \draw[-latex] (sample1) -- (encoder);
    \node[basestyle, align=center, text width=1.5cm, right= 0.5cm of markov, dashed] (logit_pred) {sampled logits $\hat{z}_i$};
    \node[basestyle, text width=1.25cm, align=center, right= 0.5cm of logit_pred] (mean) {Softmax + average};
    \node[basestyle,, align=center, right= 0.5cm of mean, dashed] (pred) {$\hat{p}_i$};

    % \draw[-latex] (-1, 4) -- (sample1);

    \draw[-latex] (encoder) -- (-0.15, 4) -- (-0.15, 4.95) -- (model);
    \draw[-latex] (encoder) -- (-0.15, 4) -- (-0.15, 3) -- (model_u);

    \draw[-latex, color=red] (model) -- (logit);
    \draw[-latex, color=teal] (model_u) -- (sigma);

    \draw[-latex, color=red] (logit) -- (7, 4.95) -- (markov);
    \draw[-latex, color=teal] (sigma) -- (7, 3.05) -- (markov);

    \draw[-latex] (markov) -- (logit_pred);
    \draw[-latex] (logit_pred) -- (mean);
    \draw[-latex] (mean) -- (pred);

    \begin{pgfonlayer}{background}
        \node[category box, fill=modelColor!5, draw=modelColor, fit=(models)] {};
    \end{pgfonlayer}

    \draw[-latex, dashed] (pred) -- (14.05, 6.25) -- (1.5, 6.25) -- (model);
    \node[] at (12.4, 6) {Cross-entropy loss};
    \draw[-latex, dashed] (pred) -- (14.05, 2) -- (1.5, 2) -- (model_u);
    % \node[] at (7.5, 1.75) {stochastic log-likelihood loss};

    \draw[-latex, dashed] (pred) -- (14.05, 6.25) -- (-1.25, 6.25) -- (encoder);
    \draw[-latex, dashed] (pred) -- (14.05, 2) -- (-1.25, 2) -- (encoder);

\end{tikzpicture}}
        \caption{The first step of AUC consists of training a single sample uncertainty-aware prediction network $M_u$.
            The network is trained to predict the gesture class and the aleatoric uncertainty of the prediction, and Monte Carlo integration is used to obtain uncertainty-weighted logits.
        }
        \label{fig:method:step1}
        \vspace{0.5cm}
    \end{subfigure}
    \begin{subfigure}[t]{\textwidth}
        \centering
        \resizebox{0.8\textwidth}{!}{\input{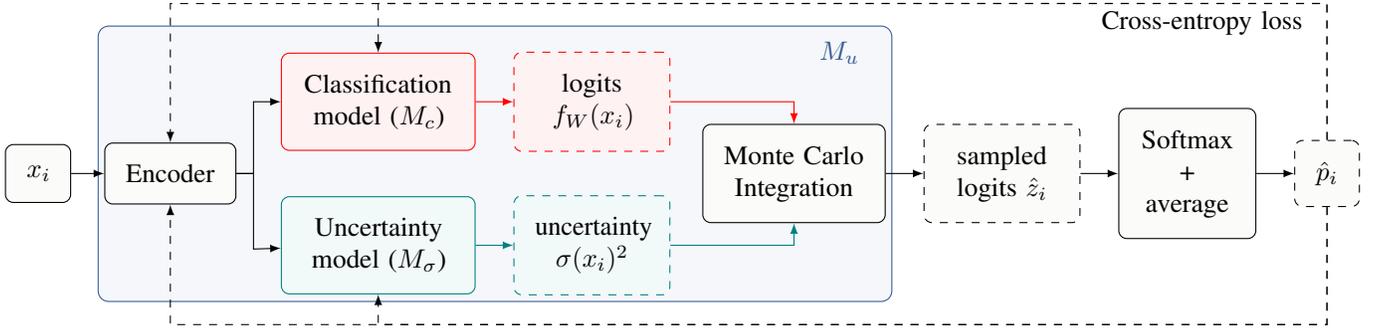}
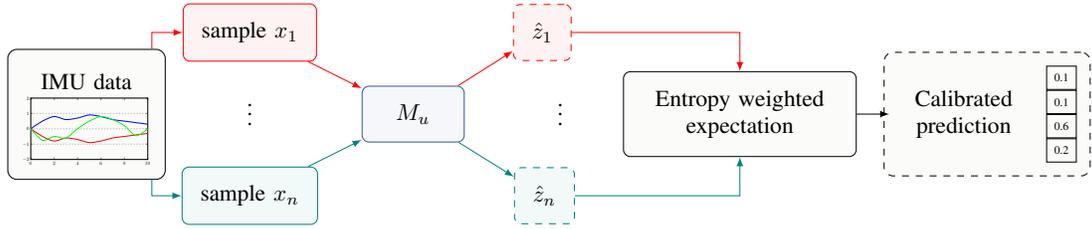}
        \caption{Once $M_u$ is trained, multiple samples of IMU data are used to obtain multiple predictions.
            The final prediction is the entropy-weighted expectation from those samples.
        }
        \label{fig:method:step2}
    \end{subfigure}
    \caption{Two-step pipeline for entropy-weighted gesture detection.
        In the first step (\cref{fig:method:step1}), an uncertainty-aware classification network $M_u$ is trained on a single sample of IMU data.
        $M_u$ is then used (\cref{fig:method:step2}) to obtain predictions from multiple samples and the final prediction is the entropy-weighted prediction from those samples.
    }
    \label{fig:method:entropy-agg-vis}
\end{figure*}

In this section, we describe UAC---uncertainty-aware calibration---a method for calibrated gesture recognition from IMU data tailored for both in and out of distribution scenarios.
UAC is used in a two steps process: first, an uncertainty-aware prediction network $M_u$ is trained on single samples of IMU data, collected from multiple users---see the top part of \cref{fig:intro:image}.
\cref{sec:method:subsec:uncertainty_classifier} presents how the network $M_u$ simultaneously learns to perform classification on samples of the IMU data while also learning the epistemic uncertainty of the model and uses both to calibrate its predictions.
In the second step presented in \cref{subsec:method:multi_sample_uncertainty_weighted_prediction}---see the bottom part of \cref{fig:intro:image}---$M_u$ is used to obtain predictions from multiple samples of IMU data from the same gesture, improving accuracy by leveraging the entropy of uncertainty-aware predictions.
% Our strategy removes the need for a post-processing step using a validation set, such as temperature scaling~\cite{pmlr-v70-guo17a}.

The pipeline of the proposed approach is described in \cref{fig:method:entropy-agg-vis}.
% \begin{enumerate}
%     \item The gesture recognition base model takes the segmented instances as inputs and outputs the predicted probabilities of each input.
%     \item Apply the entropy calibration: multiply each prediction probability with the entropy scalar $s$.
%     \item Aggregate all the instances of one gesture file to get the summed-up predicted probability and take the softmax.
% \end{enumerate}

\subsection{IMU Data Pre-Processing}
\label{sec:method:problem-definition}

Our input consists of a set $S$ of multiple sequences $s$ of IMU data, each corresponding to a specific gesture.
For each sequence $s \in S$, a sliding window is used to extract a set of $N$ samples $X_s = \{x_i\}$, with a fixed stride length and where each sample $x_i$ is of fixed size $m$. Let $X = \bigcup_{s\in S} X_s$ represent the set of all samples $x_i$ extracted from all sequences in $S$.
The samples $x_i$ are normalized using:
\begin{equation}
    x_i = \frac{x_i - \mu}{\sigma}
\end{equation}
where $\mu$ and $\sigma$ are the mean and standard deviation of all values in $X$.

It should be noted that when running experiments $\mu$ and $\sigma$ are calculated using only the training dataset.
Thus, the normalization is not impacted by unseen data, and no information from the test dataset is leaked to the network at training time---which is especially important for validation of our method in OOD scenarios.

% For each sample $x_i \in X$, we train a classification model $f(x_i)$ in a supervised manner to predict the gesture's class corresponding to each sample IMU data.
% It should be noted that our method is independent on the type of model $f$ trained as long as the model returns a set of probabilities---in \cref{sec:eval} we show improved classification results and calibration with both CNN and LSTM CNN models.

\subsection{Epistemic Uncertainty Classifier}
\label{sec:method:subsec:uncertainty_classifier}

The first step of UAC consists of training (in a supervised manner) a classifier $M_u$ that predicts a calibrated class of a given sample from uncalibrated logits and predicted epistemic uncertainty of the model.
Given the set of normalized samples obtained in the previous section, the objective of our method is to predict the probability distribution $P(y|x_i)$ of the gesture class $y$ given the set of samples $X$.

% Calibration of the classifier is crucial for use of AI in safety critical application.
Since, as seen in \cref{sec:rw}, calibration and uncertainty estimation are related, our hypothesis is that \emph{uncertainty prediction can be learned and used to improve calibration} of the classifier.
% , especially in low-confidence region of OOD data.
Since IMU data is noisier and less descriptive than other sensor modalities such as images, to improve the robustness of the uncertainty prediction, we propose to predict the uncertainty over \emph{a feature space} estimated from an encoder trained over the IMU data, instead of the raw IMU data.
Predicted uncertainties are then used to model the epistemic uncertainty of $M_u$, by modeling the weights as distributions.
For a flowchart of the uncertainty-aware prediction method, see \cref{fig:method:step1}.
% instead of look at the aleatoric uncertainty---i.e. uncertainty that stem from the data.
% and to predict and capture the epistemic uncertainty of the model alone instead of look at the aleatoric uncertainty---i.e. uncertainty that stem from the data---we propose to
% Furthermore, we propose to \emph{predict the uncertainty over a feature space estimated from an encoder} trained over the IMU data, instead of the raw IMU data to improve the robustness of the uncertainty prediction.

% $M_u$ predicts both the gesture's class corresponding to each sample IMU data $x_i$, and its associated uncertainty $\sigma_i^2$, then uses the uncertainty prediction to calibrate the classifier's prediction.
$M_u$ is composed of a 1D CNN encoder network that converts the sample input $x_i$ into a set of features $h_i$ from a feature space $h(x_i)$.
Using the features $h_i$ as inputs, two distinct networks ($M_c$ and $M_\sigma$) are trained to respectively predict the class logits $f_W(h_i)$ and the uncertainty $\sigma^2$ associated with $h_i$.
% While the classifier architecture $M_c$ predicts the class logits $f_W(f_i)$, $M_\sigma$ estimates the uncertainty $\sigma^2$ associated with the $f_i$ prediction.
We empirically observed that predicting the uncertainty from a feature space into which the IMU data is transformed, rather than using the IMU data directly as input, improves the robustness of the uncertainty prediction.
Indeed, the encoder acts as a denoiser and the feature space in which the IMU is projected is less noisy and more descriptive than the raw IMU data, allowing the uncertainty prediction to converge.

%It should be noted that our method is independent on the type of backbone model trained as long as it returns a set of probabilities---in \cref{sec:eval} we show improved classification results and calibration with both CNN and LSTM CNN models.

% Inspired by the work of \textcite{kendall} on pixel-wise classification of image, we extend the classification network to explicitely model aleatoric uncertainty.
% , following the work of \textcite{kendall}, who applied this idea to image data for pixel-wise classification tasks.
% We adapt the model of aleatoric uncertainty to the classification of IMU data.

% To take into account the variance of the model's weights, the predicted uncertainty $\sigma^2$ is used to model a Gaussian distribution that represents the uncertainty of the predicted logits.
To model the epistemic uncertainty of the model over the prediction $f_W(x_i)$, we represent the uncertainty over the model's weight for each input sample $x_i$ as a Gaussian distribution with mean $f_W(x_i)$ and variance $\sigma^2(x_i)$:
\begin{equation}
    \hat{z}_i | W \sim \mathcal{N}(f_W(h_i), (\sigma(h_i))^2)
    \label{eq:uncertainty_model}
\end{equation}
where $f_W(h_i)$ represents the predicted logits for input $x_i$, and $\sigma(h_i)$ its predicted uncertainty.
Since the integral does not have a closed-form solution, we approximate the expectation using Monte Carlo integration to sample $T$ candidate logits $\hat{z}_i$ on the distribution represented in \cref{eq:uncertainty_model}---in our experiments, $T$ is experimentally set to 100.
It should be noted that each logit is sampled independently from the Gaussian distribution, which means that the uncertainty is modeled as independent and identically distributed (i.i.d.) across the logits.

The softmax function is then applied to each $\hat{z}_i$ to obtain the predicted class probabilities:
\begin{equation}
    \hat{p}_{i,c} = \frac{\exp(\hat{z}_{i,c})}{\sum_{k=1}^{C} \exp(\hat{z}_{i,k})}
    \label{eq:softmax}
\end{equation}
where $C$ is the number of classes and $\hat{z}_{i,c}$ is the logit for class $c$ for input $x_i$.
The final uncertainty-aware prediction is the mean of the predicted class probabilities over the $T$ samples:
\begin{equation}
    \hat{p}_i = \frac{1}{T} \sum_{t=1}^{T} \hat{p}_{i,t}
\end{equation}

To train $M_c$ and $M_\sigma$, the cross-entropy loss $\mathcal{L}_c$  is used:
\begin{equation}
    \mathcal{L}_{c} = -\sum_{i=1}^{N} y_{i} \log(\hat{p}_{i})
    \label{eq:method:cross_entropy_loss}
\end{equation}
%
% where $T$ is the number of Monte Carlo samples, $c$ is the true class and $\hat{z}_{i,t,c}$ is the $c$-th logit of the $t$-th sample.
When uncertainty is high, the sampled logits $\hat{z}_{i,t}$ will vary significantly across samples, causing the average probability to spread across different classes and reducing the log probability $\log p$ of the correct class.
As a result, the gradient of the loss with respect to $\hat{z}_i$ becomes smaller, and high-uncertainty data points contribute less to the total loss.
Conversely, when the uncertainty is low, logits $\hat{z}_{i,t}$ stay close to $f_W(x_i)$ for all samples.
The predicted probabilities are more confident and concentrated on one class, the loss for these points is not attenuated, and they contribute fully to the loss.
This strategy encourages the model to focus on reliable, confident predictions while reducing the impact of noisy, uncertain data points.

The cross-entropy loss is used with the classification network $M_c$ and the variance estimator $M_\sigma$ to respectively learn the class logits and the variance $\sigma^2$.
Inspired by the work of \textcite{kendall}, we improve numerical stability by training the model to predict the log-variance $s_i = \log \sigma_i^2$ instead of the variance ensuring that the variance remains positive.

% Hence, we compute the expected log-likelihood of the predicted probabilities given by:
% \begin{equation}
%     \log \mathbf{E}_{\hat{z}_i \sim \mathcal{N}(f_W(x_i), (\sigma_W(x_i))^2)}\left[\hat{p}_{i,c}\right]
% \end{equation}
% where $\hat{p}_{i,c}$ is the probability of the true class $c$.

\subsection{Multi-Sample Entropy-Weighted Prediction}
\label{subsec:method:multi_sample_uncertainty_weighted_prediction}

Once the $M_u$ classification network is trained, it can be used to estimate the uncertainty---and thus the calibrated probabilities---of each sample in the input IMU data of a gesture.
% However, gesture and uncertainty estimation can still be sensitive to sensor noise and similarity between motions, especially on out-of-distribution IMU sequences.

In \cref{sec:method:problem-definition}, it is shown how the samples are generated from a given IMU sequence through a moving window of fixed length to generate the samples.
Such a strategy means that the samples are not equally representative of the motion---e.g. some samples might correspond to the start of the motion, while others the end of the motion, or even a moment where no motion is registered.
To alleviate this issue, previous work~\cite{mollyn2022samosa} propose to sum the predictions of multiple samples and select the class with the highest summed prediction as the final output, increasing accuracy.
However, in this case, the model's output is no longer a set of probabilities.
A simple solution to obtain a probability distribution would be to average the predictions across all samples. However, we observed that this strategy leads to slightly better accuracy at the cost of network calibration.
% we can use the summed prediction as the input to the Softmax function to compute the final prediction;

To improve gesture predictions while maintaining, or improving, network calibration, we propose a novel multi-sample entropy-weighted prediction strategy---see \cref{fig:method:step2}.
Using multiple samples from an input IMU sequence of a gesture and the entropy of each sample's prediction as a weight, the final prediction is computed as the expectation of all sample predictions.

The entropy of a sample is the negative of the sum of the product of the predicted probabilities and their log:
\begin{equation}
    \label{eq:method:entropy}
    H_i = -\sum_{c=1}^{C} \hat{p}_{i,c} \log \hat{p}_{i,c}
\end{equation}
A high entropy indicates that the predicted probabilities are spread across multiple classes, while a low entropy indicates that the predicted probabilities are concentrated on one class.
Hence, samples with a high entropy should have a lower weight in the final prediction, while samples with a low entropy should have a higher weight.

Given a set of $K$ samples, $X_s = \{x_i\}$, from the input IMU data of a gesture, we first estimate the uncertainty-aware logits $\hat{p}_i$ of each sample.
The multi-sample entropy-weighted prediction is defined as the expectation of the probabilities, weighted by the entropy of each sample.
To be able to use the entropy as a weight where high entropy means low impact, and low entropy means high impact, entropy values are rescaled.
Since entropy values range between $0$ and $log(C)$, we rescale the entropy value to a new measure $W_i$ as follows:
\begin{equation} \label{eq:entscalar-def}
    W_i = \frac{log(C) - H(x_i)}{log(C)}
\end{equation}
where $W_i$  between $0$ and $1$. A value of $0$ corresponds to the highest entropy, and $1$ corresponds to the lowest entropy.

For a sequence of IMU data $r$ comprised of $K$ samples $x_i$, the final prediction is computed as the expectation of individual sample predictions:
\begin{equation}
    E(r) = \frac{1}{K}\sum_{i=0}^K W_i \hat{p}_i
\end{equation}

\section{Experiments and Implementation}
\label{sec:eval}

In this section, we present the metrics and the datasets used for the evaluation, as well as the implementation details and experimental setup.

\subsection{Metrics}

In this paper, we introduce a method aimed at improving the accuracy and calibration of gesture detection models in OOD scenarios.
To assess both the accuracy and calibration of the model, we employ three key metrics.

The \emph{accuracy} is assessed against the ground truth labels using the following formula:
\begin{equation}
    accuracy = \frac{1}{N} \sum_{i=1}^N \mathbf{1}(y_i = \hat{y}_i)
\end{equation}
where $y$ is the target label and $\hat{y}$ is the predicted label.

Conversely, the calibration is evaluated using the \emph{Expected Calibration Error} (ECE) and the \emph{Negative-Log-Likelihood} (NLL) \cite{pmlr-v70-guo17a}.
The ECE quantifies the model's calibration by comparing its predicted confidence with its accuracy.
This is achieved by dividing the probabilities into equally sized bins and calculating the absolute difference between accuracy and confidence for each bin:
\begin{equation}
    ECE = \sum_{m=1}^M \frac{|B_m|}{n} |acc(B_m)- conf(B_m)|
\end{equation}
A well-calibrated model should assign high confidence to correct predictions and low confidence to incorrect ones.
If the model is overconfident, the ECE will be high, indicating poor calibration.
% Thus, the ECE measures how much a model's predicted confidence differs from its actual accuracy---i.e. a well-calibrated model should be confident only when it is correct.
On the other hand, the NLL measures how well the predicted probabilities generated by the model align with the true probabilities of the outcomes.
It is expressed as:
\begin{equation}
    NLL = -\frac{1}{N} \sum_{i=0}^N log(p_{i, y_i})
\end{equation}
with $p_{i, y_i}$ the probability assigned to the true class $y_i$ of sample $i$ and $N$ the total number of samples.

\subsection{Dataset}

To evaluate the robustness of our method, we conducted experiments using three publicly available datasets for gesture recognition based on IMU data.
Below, we provide a brief overview of each dataset; for detailed information, please refer to their respective papers.

\subsubsection{Wireless Sensor Data Mining (Wisdm)} This dataset~\cite{wisdm} includes data collected from 51 participants who performed 18 different activities, each lasting 3 minutes, using both a smartphone and a smartwatch (LG G Watch).
The dataset includes accelerometer and gyroscope data collected at 20~Hz from both devices, totaling four sensors.
In our paper, we focus solely on smartwatch sensor data.
For each subject and gesture, the accelerometer and gyroscope data collected are provided in separate files, and each sample has an associated timestamp.
% Data points with the same timestamp represent the same moment in time.
We combine the accelerometer and gyroscope samples based on matching timestamps, resulting in data points with 6 features (3 from each sensor: x, y, z accelerometer, and gyroscope readings).
Activities cover various daily tasks such as walking, eating, and typing.

\subsubsection{Samosa Dataset} This dataset~\cite{mollyn2022samosa} contains 9 dimensional IMU data collected from 20 participants performing daily activities---acceleration, rotation velocity, and orientation recorded by a smartwatch on each participant's wrists.
The dataset covers 26 daily activities, including common arm and hand movements such as clapping, drinking water, and washing hands.

\subsubsection{The University of Southern California Human Activity Dataset (USCHAD)} This dataset~\cite{uschaddataset} contains IMU data---accelerometers and gyroscope---placed on the front right hip of the 14 participants recording 12 common daily, such as walking forward, jumping, standing, and sleeping.

% \subsubsection{UCI-Human-Activity-Recognition (UCIHAR)~\cite{Anguita2013APD}} contains recordings from 30 participants aged between 19 and 48 years.
% Each participant wore a Samsung Galaxy S II smartphone at their waist while performing six common activities: standing, sitting, laying down, walking, walking upstairs, and walking downstairs.
% The smartphone’s accelerometer and gyroscope recorded 3-axis linear acceleration and angular velocity data at a sampling rate of 50 Hz, which was preprocessed using noise reduction filters. The data was split into fixed-width sliding windows of 2.56 seconds with 50\% overlap.

When evaluating OOD scenarios, we divide the dataset into training, validation, and test sets based on subject IDs so that each subject appears in only one unique set.
This approach guarantees that the model is trained on data from a specific group of subjects and tested on an entirely new set of subjects.
When evaluating in-distribution scenarios, the split is done so that subjects are present in all data sets.

The data is divided into a training set $X_{train}$, a validation set $X_{val}$, and a test set $X_{test}$ with a ratio of $60:20:20$.
During the normalization step outlined in \cref{sec:method:problem-definition}, it should be noted that while $X_{test}$ is normalized, it is not used to compute the mean and standard deviation to ensure that the model does not benefit from any information in the test set.
Additionally, we employ a stride length of 10 data points for our analysis.

\subsection{Implementation Details}
\label{subsec:implementation}

% As presented in \cref{sec:method},
$M_u$'s encoder (see \cref{fig:method:step1}) consists of three 1D convolutional layers followed by a ReLU activation and batch normalization.
The first, second, and third convolutional layers have 128, 128, and 256 output channels with a kernel size of 10 and a stride of 1 respectively.
The second and third convolutional layers are followed by dropouts with a rate of 0.25 and max-pooling with a kernel size of 2.
The output from the encoder is flattened before being inputted in the classification model $M_c$ which consists of two fully connected layers, with the first having 256 units and the second $K$ units (with $K$ being the number of classes).
After the first fully connected layer, we apply a dropout rate of $0.5$.

The hyperparameters (learning rate, dropout rate, batch size) and model configurations (e.g., number of units per layer, pooling layers, batch normalization) were selected based on the performance metrics from the validation split and found using grid search.
% our method is independent of the classification model $M_c$ (see \cref{fig:method:entropy-agg-vis} used as long as it predicts probabilities given IMU data as input.
% Hence, two different architectures are used for the tests, a CNN-based architecture (referred to as $M_c^{cnn}$ in this paper) and a long-short term memory (LSTM) CNN network (referred to as $M_c^{lstm}$ in this paper)---see \cref{sec:exp:fig:networks} for a detailled representation of both networks.
% Both networks are trained using the cross-entropy loss (\(L_{\text{CE}}\)).
During training, we use the Adam optimizer with a batch size of $64$.
The learning rate is initially set at $1\times10^{-6}$ and is reduced by a factor of 0.1 if the accuracy of the validation does not improve after 10 epochs.
% On the other hand, $M_c^{lstm}$ consists of three 1D convolutional layers followed by two LSTM layers of 128 hidden dimensions each.
% To obtain the predicted probabilities, the LSTM outputs are passed through two fully connected layers and a softmax layer.
The uncertainty network $M_\sigma$ is a 2-layer multilayer perceptron (MLP) that predicts the log variance of $M_c$'predictions.
% We train the classification network and the uncertainty network jointly using the loss function described in \cref{eq:method:aleatoric_loss}.
% The uncertainty prediction models are referred to as $M_u^{cnn}$ and $M_u^{lstm}$ in this paper.

% Our implementation (including models and experiments) can be found online.\footnote{Link will be added once accepted for publication}

\subsection{Baseline Methods}
\label{sec:eval:baseline}

To evaluate our uncertainty estimation strategy, we implemented three baseline methods derived from state-of-the-art calibration techniques, which we use as substitutes for $M_u$ in our experiments.
The three designs are outlined below:

% \subsubsection{Baseline CNN Model}

% The baseline classification model is a CNN that has the same architecture as the encoder from \ref{subsec:implementation} followed by two fully connected layers. However, it is trained using the standard cross-entropy loss without uncertainty estimation.

\subsubsection{Entropy-Maximization (EM) for Misclassified Samples}
\label{sec:method:maximization}

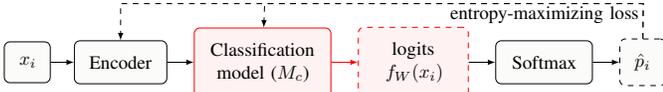
\begin{figure}[t]
    \centering
    \resizebox{0.49\textwidth}{!}{\begin{tikzpicture}

    % \begin{scope}[local bounding box=imu]
    %     \node[base node] (data) at (-5, 4.5) {IMU data};
    %     \node at (-5, 3.75) {\includegraphics[width=2cm]{sections/imgs/IMU_graph.pdf}};
    % \end{scope}
    \node[basestyle, text width=0.3cm, align=center] (sample1) at (-3, 4) {$x_i$};
    % \draw[-latex] (-3.75, 4) -- (sample1);

    \begin{scope}[local bounding box=models]

        \node[basestyle, align=center] (encoder) at (-1.25, 4) {Encoder};
        \node[samples1, text width=2cm, align=center, right= 0.5cm of encoder] (model) {Classification model ($M_c$)};
        \node[samples1, text width=1.5cm, align=center, right= 0.5cm of model, dashed] (logit) {logits $f_W(x_i)$};
        % \node[text=modelColor] at (5, 5) {EM};
    \end{scope}

    \draw[-latex] (sample1) -- (encoder);
    \node[basestyle, text width=1.25cm, align=center, right= 0.5cm of logit] (softmax) {Softmax};
    \node[basestyle,, align=center, right= 0.5cm of softmax, dashed] (pred) {$\hat{p}_i$};

    \draw[-latex] (encoder) -- (-0.15, 4) -- (-0.15, 4) -- (model);
    \draw[-latex, color=red] (model) -- (logit);
    \draw[-latex] (logit) -- (softmax);
    \draw[-latex] (softmax) -- (pred);

    \begin{pgfonlayer}{background}
        % \node[category box, fill=modelColor!5, draw=modelColor, fit=(models)] {};
        % \node[category box, fill=backgroundColor!5, draw=black, fit=(imu)] {};
    \end{pgfonlayer}

    \draw[-latex, dashed] (pred) -- (8.55, 5.1) -- (1.45, 5.1) -- (model);
    \node[] at (6.7, 4.9) {entropy-maximizing loss};

    \draw[-latex, dashed] (pred) -- (8.55, 5.1) -- (-1.25, 5.1) -- (encoder);

\end{tikzpicture}}
    \caption{The Entropy Maximization (EM)~\cite{max-entropy} baseline model is trained to maximize the entropy of wrong predictions through an entropy-maximization loss.
    }
    \label{fig:experiments:EM}

\end{figure}

As seen previously, entropy can act as a measure of prediction uncertainty.
We implement a variation of the method proposed by \textcite{max-entropy}---a method to maximize the entropy of incorrect predictions--- by training the encoder and $M_c$ network to maximize the entropy of incorrect predictions---see \cref{fig:experiments:EM}.
% Entropy weighted aggregation prioritizes more confident predictions, however in this case overconfident wrong predictions can skew the aggregation towards the wrong class.
In our implementation, we train the same encoder and $M_c$ network as in \cref{subsec:implementation} with a softmax function instead of Monte Carlo integration to classify gesture using the entropy-maximizing loss:
% When incorrect predictions have higher entropy, they carry less weight during the aggregation process ensuring that the final gesture classification is driven more by the confident correct predictions.
%An entropy maximization term for misclassified samples (maximizing entropy equivalent to minimizing -entropy).
\begin{equation}
    \begin{split}
        \mathcal{L}\left(x_i\right) = &
        \mathcal{L}_{CE}\left(p\left(y|x_i\right), y\right) \\
        & + \lambda . I_{m}\left(x_i\right) . H\left(s\right)
    \end{split}
\end{equation}
where, given a sample $x_i$ with label $y$, $\mathcal{L}_{CE}$ represents the cross-entropy loss, $H\left(p\left(y|x_i\right)\right)$ is the entropy of the predicted distribution, and $I_{m}\left(x_i\right)$ is an indicator function for whether the sample is misclassified.
$\lambda$ is a parameter that controls the influence of the entropy maximization term.
The lambda coefficient was tuned using a grid search over the predefined set of values \(\{0.1, 0.15, 0.2, 0.25, 0.4, 0.5, 0.6, 0.8, 0.9\}\).
Given that \(\lambda = 0.2\) yielded the best results across all datasets,  this value was selected for our experiments.
% This encourages the model to express more uncertainty on wrong predictions which leads to improved aggregation performance and calibration.

\subsubsection{Temperature Scaling}
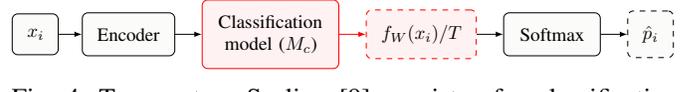
\begin{figure}[t]
    \centering
    \resizebox{0.49\textwidth}{!}{\begin{tikzpicture}

    \node[basestyle, text width=0.3cm, align=center] (sample1) at (-3, 4) {$x_i$};

    \begin{scope}[local bounding box=models]
        \node[basestyle, align=center] (encoder) at (-1.25, 4) {Encoder};
        \node[samples1, text width=2cm, align=center, right= 0.5cm of encoder] (model) {Classification model ($M_c$)};
        \node[samples1, text width=1.5cm, align=center, right= 0.5cm of model, dashed] (logit) {$f_W(x_i) / T$};
    \end{scope}

    \draw[-latex] (sample1) -- (encoder);
    \node[basestyle, text width=1.25cm, align=center, right= 0.5cm of logit] (softmax) {Softmax};
    \node[basestyle,, align=center, right= 0.5cm of softmax, dashed] (pred) {$\hat{p}_i$};

    \draw[-latex] (encoder) -- (-0.15, 4) -- (-0.15, 4) -- (model);
    \draw[-latex, color=red] (model) -- (logit);
    \draw[-latex] (logit) -- (softmax);
    \draw[-latex] (softmax) -- (pred);

    \begin{pgfonlayer}{background}
        % \node[category box, fill=modelColor!5, draw=modelColor, fit=(models)] {};
    \end{pgfonlayer}

\end{tikzpicture}}
    \caption{Temperature Scaling~\cite{pmlr-v70-guo17a} consists of a classification model $M_c$ where the logits are scaled by a factor $T$.
        $T$ is learned after training $M_c$, to improve the calibration of the model on a validation set.
    }
    \label{fig:experiments:temp_scaling}

\end{figure}
Temperature scaling~\cite{pmlr-v70-guo17a} is a post-processing method that adjusts the confidence scores of the predicted probabilities of a model.
The calibrated probabilities are computed as:
\[
    \hat{\mathbf{p}} = \text{softmax}\left(\frac{\mathbf{z}}{T}\right),
\]
where \(\mathbf{z}\) represents the logits, and $T$ is the temperature parameter.
% By scaling the logits with $T$, temperature scaling adjusts the confidence of the model’s predictions.
The temperature parameter $T$ is optimized by minimizing the Negative Log-Likelihood (NLL) on a separate validation set. The NLL is computed as:
\[
    L_{\text{cal}}(T) = -\frac{1}{N} \sum_{i=1}^{N} \log \left( \hat{p}_{i, y_i} \right),
\]
where \(\hat{p}_{i, y_i}\) is the calibrated probability for the true class $y_i$ of sample $i$, and $N$ is the total number of validation samples.
If $T=1$, there is no scaling, and the probabilities remain unchanged.
If $T>1$, the model’s confidence is reduced, making the output probabilities more uniform.
Conversely, for $T<1$, the model’s confidence is increased, leading to sharper probability distributions.

The temperature parameter $T$ is optimized using a separate validation set.
To ensure that $T$ remains positive, we optimize $\log T$.
Additionally, we constrain $T$ to be less than 5 to prevent excessively large values that could result in overly flattened probability distributions.
Once the optimal temperature parameter has been learned, logits are scaled by $T$ during evaluation to produce calibrated probabilities.
In this paper, models with temperature scaling applied are referred to as ``temp scaling''.

In the experiments, we kept the encoder and $M_c$ network with the same representation as in \cref{sec:method}.
See \cref{fig:experiments:temp_scaling} for a graphic representation of the temperature scaling method and models.

\subsubsection{Last-Layer Bayesian Neural Network with Laplacian approximation}

\begin{figure}[t]
    \centering
    \resizebox{0.49\textwidth}{!}{\input{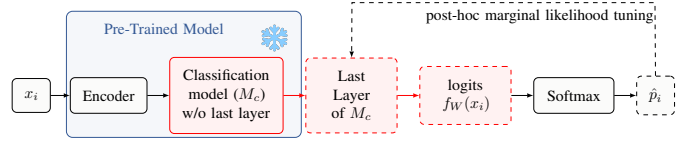}}
    \caption{Implementation of the last-layer Bayesian Neural Network with Laplacian approximation~\cite{NEURIPS2021_a7c95857} baseline used in the experiments.
        The pre-trained model is assumed to be a feature map and the Laplace-approximated posterior is calculated for the last layer only.}
    \label{fig:experiments:bayesian}
\end{figure}

In Bayesian neural networks (BNNs), prediction uncertainty is estimated by marginalizing over the posterior distribution of the network's weights:

\begin{equation} \label{eq:pred_uncertainty}
    p (y|x, D) = \int_{\theta}p(y|x, \theta') p(\theta'|D) d\theta'
\end{equation}

However, this integral becomes intractable for deep models.
To address this, \textcite{NEURIPS2021_a7c95857} propose to use the Laplace approximation to estimate the posterior distribution during Bayesian inference.
Thus, the posterior $p(\theta|\mathcal{D})$ is approximated by a Gaussian distribution centered at the Maximum A Posteriori (MAP) estimate of the weights, denoted as \textbf{$\theta_{MAP}$}.
The covariance matrix is approximated using the inverse of the negative Hessian of the log-posterior evaluated at $\theta_{MAP}$, yielding the Gaussian distribution:
\begin{align}
    \label{eq:gauss_posterior}
    p(\theta|\mathcal{D}) & \approx \mathcal{N}(\theta; \theta_{MAP}, \Sigma)                                        \\
    \text{with} \; \Sigma & = -\left(\nabla_{\theta}^2 \mathcal{L}(\mathcal{D}; \theta) |_{\theta_{MAP}}\right)^{-1}
\end{align}

In this paper, we implement the Laplace approximation as a post-hoc method to estimate prediction uncertainty efficiently by tuning the last layer of a classification model to be Bayesian--similar to the method proposed by \textcite{snoek2015scalable}.
First, we train the same encoder and $M_c$ network in a supervised manner as in \cref{subsec:implementation}.
We then freeze all layers except for the last one which is a fully connected layer that outputs the gesture class prediction, treating the output of the penultimate layer as fixed features---i.e. only the last layer’s weights are modeled as random variables.
Then, we apply the approximation to the last layer of the CNN to approximate the posterior distribution of the weights.
% We use the Laplace-PyTorch library to apply the approximation to the last layer of the CNN model and estimate the posterior distribution of the weights.
% A key advantage of the Laplace-Pytorch library is that it incorporates all the requisite mathematical calculations for approximating the posterior distribution, thereby eliminating the need for users to perform these calculations themselves.
% The pre-trained CNN model is passed as a parameter to the Laplace class, which then fits the last layer’s weights using the training dataset and computes the posterior distribution.

Bayesian Laplacian models are referred to as "Bayesian" and a representation of the method is shown in \cref{fig:experiments:bayesian}.
\section{Discussion}
\label{sec:discussion}

In this section, we present the experimental results for both in-distribution and out-of-distribution scenarios.
The primary distinction between these scenarios lies in the composition of the training, validation, and test sets.
The out-of-distribution test set includes subjects who are not present in the training and validation sets, making classification more challenging and increasing prediction uncertainty.
In contrast, in the in-distribution scenario, all sets (training, validation, and test) contain mixed IMU data from all subjects.
This methodology enables the evaluation of general model performance through the in-distribution scenario, as well as the model's robustness in a more realistic out-of-distribution scenario, offering insights into the method's accuracy and calibration under domain shift.

\subsection{Comparison to Baseline Methods}

In this section we present the results of the experiments in both out and in distribution scenarios, comparing UAC against the baseline methods outlined in \cref{sec:eval:baseline}---temperature scaling, entropy maximization, and Bayesian Laplacian network.

\subsubsection{Out-of-distribution Scenario}

\begin{table}[t]
    \centering
    \caption{Comparison of UAC (ours) to the baseline calibration methods entropy maximization (EM), temperature scaling, and Bayesian network, each followed by entropy weighted expectation to get the final prediction on the Wisdm dataset in out-of-distribution scenario. (The best value for each metric is in bold, and the second best value is underlined).}
    \label{tab:cnn-sw-wisdm-baselines}
    \resizebox{\columnwidth}{!}{%
        \begin{tabular}{lccc}
            \toprule
            Method       & Accuracy$\uparrow$          & ECE$\downarrow$               & NLL$\downarrow$               \\
            \midrule
            % Classification & - & -  & - \\
            EM           & $0.58 \pm 0.07$             & $0.164 \pm 0.054$             & \underline{$1.275 \pm 0.306$} \\
            Temp Scaling & \underline{$0.64 \pm 0.07$} & \underline{$0.157 \pm 0.049$} & $1.367 \pm 0.344$             \\
            Bayesian     & \underline{$0.64 \pm 0.07$} & $0.544 \pm 0.058$             & $2.492 \pm 0.099$             \\
            % UAC (no $\sigma$) & $0.63 \pm 0.06$ &  \underline{$0.140 \pm 0.071$} & $1.321 \pm 0.311$ \\
            UAC (ours)   & $\mathbf{0.75 \pm 0.09}$    & $\mathbf{0.103 \pm 0.027}$    & $\mathbf{1.098 \pm 0.569}$    \\

            \bottomrule
        \end{tabular}
    }
\end{table}

\begin{table}[t]
    \centering
    \caption{Comparison of UAC (ours) to the baseline calibration methods entropy maximization (EM), temperature scaling, and Bayesian network, each followed by entropy weighted expectation to get the final prediction on the Samosa dataset in out-of-distribution scenario. (The best value for each metric is in bold, and the second best value is underlined).}
    \label{tab:cnn-sw-samosa-baselines}
    \resizebox{\columnwidth}{!}{%
        \begin{tabular}{lccc}
            \toprule
            Method       & Accuracy$\uparrow$          & ECE$\downarrow$               & NLL$\downarrow$               \\
            \midrule
            % Classification & - & -  & - \\
            EM           & \underline{$0.47 \pm 0.04$} & \underline{$0.128 \pm 0.054$} & \underline{$1.820 \pm 0.136$} \\
            Temp scaling & \underline{$0.47 \pm 0.05$} & $0.138 \pm 0.044$             & $1.854 \pm 0.132$             \\
            Bayesian     & \underline{$0.47 \pm 0.04$} & $0.408 \pm 0.037$             & $3.016 \pm 0.031$             \\
            % UAC (no $\sigma$) &  \underline{$0.47 \pm 0.04$} &  \underline{$0.100 \pm 0.045$} &  \underline{$1.778 \pm 0.156$} \\
            UAC (ours)   & $\mathbf{0.51 \pm 0.04}$    & $\mathbf{0.063 \pm 0.025}$    & $\mathbf{1.653 \pm 0.128}$    \\
            \bottomrule
        \end{tabular}
    }
\end{table}

\begin{table}[t]
    \centering
    \caption{Comparison of UAC (ours) to the baseline calibration methods entropy maximization (EM), temperature scaling, and Bayesian network, each followed by entropy weighted expectation to get the final prediction on the Uschad dataset in out-of-distribution scenario. (The best value for each metric is in bold, and the second best value is underlined).}
    \label{tab:cnn-sw-uschad-baselines}
    \resizebox{\columnwidth}{!}{%
        \begin{tabular}{lccc}
            \toprule
            Method       & Accuracy$\uparrow$          & ECE$\downarrow$               & NLL$\downarrow$               \\
            \midrule
            % Classification & - & -  & - \\
            EM           & \underline{$0.75 \pm 0.04$} & \underline{$0.114 \pm 0.044$} & $0.778 \pm 0.088$             \\
            Temp Scaling & $0.73 \pm 0.04$             & $0.126 \pm 0.048$             & $\mathbf{0.739 \pm 0.099}$    \\
            Bayesian     & $0.74 \pm 0.03$             & $0.585 \pm 0.040$             & $1.944 \pm 0.050$             \\
            % UAC (no $\sigma$) & $0.74 \pm 0.02$ & $\mathbf{0.085 \pm 0.034}$ & $\mathbf{0.682 \pm 0.107}$ \\
            UAC (ours)   & $\mathbf{0.77 \pm 0.03}$    & $\mathbf{0.090 \pm 0.044}$    & \underline{$0.746 \pm 0.256$} \\
            \bottomrule
        \end{tabular}
    }
\end{table}

Looking at the results---presented in \cref{tab:cnn-sw-wisdm-baselines} for Wisdm, \cref{tab:cnn-sw-samosa-baselines} for Samosa, and \cref{tab:cnn-sw-uschad-baselines} for Uschad---one can see that UAC consistently outperforms the baselines in terms of accuracy and model calibration across all datasets.
Specifically, on the Widsm dataset---which is the most comprehensive---UAC demonstrates an accuracy improvement of $11\%$ compared to the two second-best methods, namely temperature scaling and Bayesian NN.
For the Samosa and Uschad datasets, the accuracy improvements are $4\%$ and $2\%$ respectively compared to the second best method.
Notably, while UAC consistently achieves the highest accuracy, the second-best method varies depending on the dataset, showcasing the inconsistencies in the results of the baselines.

Looking at the calibration results, all baseline methods exhibit significantly lower calibration metrics compared to UAC.
For instance, on the Widsm dataset, UAC achieves an ECE of $0.103 \pm 0.027$ and a NLL of $1.098 \pm 0.569$.
In contrast, the second best ECE is $0.157 \pm 0.049$ for temperature scaling, and the second best NLL is $1.275 \pm 0.306$ for EM, representing an improvement of approximately $50\%$.

In conclusion, UAC demonstrates superior performance over the baselines in out-of-distribution scenarios, improving both the accuracy of prediction and model calibration.
% By accounting for prediction confidence, the uncertainty-aware approach not only improves accuracy but also improves the calibration of the model.

\subsubsection{In-Distribution Scenario}

\begin{table}[t]
    \centering
    \caption{Comparison of UAC (ours) to the baseline calibration methods entropy maximization (EM), temperature scaling, and Bayesian network, each followed by entropy weighted expectation to get the final prediction on the Wisdm dataset in in-distribution scenario. (The best value for each metric is in bold, and the second best value is underlined).}
    \label{tab:cnn-mix-wisdm-baselines-in}
    \resizebox{\columnwidth}{!}{%
        \begin{tabular}{lccc}
            \toprule
            Method       & Accuracy$\uparrow$          & ECE$\downarrow$               & NLL$\downarrow$               \\
            \midrule

            EM           & 0.82 $\pm$ 0.02             & 0.132 $\pm$ 0.022             & \underline{0.703 $\pm$ 0.099} \\
            Temp scaling & 0.78 $\pm$ 0.03             & \underline{0.127 $\pm$ 0.022} & 0.788 $\pm$ 0.092             \\
            Bayesian     & \underline{0.89 $\pm$ 0.02} & 0.716 $\pm$ 0.016             & 1.759 $\pm$ 0.016             \\
            % UAC (no $\sigma$) & 0.80 $\pm$ 0.03 & 0.082 $\pm$ 0.019 & 0.675 $\pm$ 0.077 \\
            UAC          & \textbf{0.98 $\pm$ 0.01}    & \textbf{0.057 $\pm$ 0.008}    & \textbf{0.159 $\pm$ 0.023}    \\

            \bottomrule
        \end{tabular}
    }
\end{table}

\begin{table}[t]
    \centering
    \caption{Comparison of UAC (ours) to the baseline calibration methods entropy maximization (EM), temperature scaling, and Bayesian network, each followed by entropy weighted expectation to get the final prediction on the Samosa dataset in in-distribution scenario. (The best value for each metric is in bold, and the second best value is underlined).}
    \label{tab:cnn-mix-samosa-baselines-in}
    \resizebox{\columnwidth}{!}{%
        \begin{tabular}{lccc}
            \toprule
            Method       & Accuracy$\uparrow$          & ECE$\downarrow$               & NLL$\downarrow$               \\
            \midrule
            EM           & \underline{0.77 $\pm$ 0.01} & 0.231 $\pm$ 0.009             & \underline{1.012 $\pm$ 0.035} \\
            Temp Scaling & 0.75 $\pm$ 0.02             & \underline{0.216 $\pm$ 0.014} & 1.027 $\pm$ 0.080             \\
            Bayesian     & 0.74 $\pm$ 0.02             & 0.676 $\pm$ 0.017             & 2.797 $\pm$ 0.036             \\
            % UAC (no $\sigma$) & 0.75 $\pm$ 0.02 & \underline{0.176 $\pm$ 0.015} & \underline{0.998 $\pm$ 0.071}\\
            UAC (ours)   & \textbf{0.97 $\pm$ 0.01}    & \textbf{0.113 $\pm$ 0.011}    & \textbf{0.229 $\pm$ 0.031}    \\
            \bottomrule
        \end{tabular}
    }
\end{table}

\begin{table}[t]
    \centering
    \caption{Comparison of UAC (ours) to the baseline calibration methods entropy maximization (EM), temperature scaling, and Bayesian network, each followed by entropy weighted expectation to get the final prediction on the Uschad dataset in in-distribution scenario. (The best value for each metric is in bold, and the second best value is underlined).}
    \label{tab:cnn-mix-uschad-baselines-in}
    \resizebox{\columnwidth}{!}{%
        \begin{tabular}{lccc}
            \toprule
            Method       & Accuracy$\uparrow$          & ECE$\downarrow$               & NLL$\downarrow$               \\
            \midrule
            EM           & \underline{0.89 $\pm$ 0.02} & 0.074 $\pm$ 0.007             & 0.346 $\pm$ 0.041             \\
            Temp Scaling & \underline{0.89 $\pm$ 0.02} & \underline{0.071 $\pm$ 0.012} & \underline{0.275 $\pm$ 0.045} \\
            Bayesian     & \underline{0.89 $\pm$ 0.02} & 0.716 $\pm$ 0.016             & 1.759 $\pm$ 0.016             \\
            % UAC (no $\sigma$) &\underline{0.89 $\pm$ 0.02} & \underline{0.050 $\pm$ 0.014} & \underline{0.253 $\pm$ 0.040} \\
            UAC (ours)   & \textbf{0.95 $\pm$ 0.01}    & \textbf{0.040 $\pm$ 0.002}    & \textbf{0.120 $\pm$ 0.006}    \\

            \bottomrule
        \end{tabular}
    }
\end{table}

% In addition to testing on out-of-distribution (OOD) data, it is essential to evaluate uncertainty estimation when there is no distribution shift between the training and testing data.
% For the in-distribution (ID) setting, we mix data from all the subjects and perform a standard train-validation-test split.
% This evaluation setup allows us to assess the effectiveness of the method when there is no distribution shift between the training and testing data.

In in-distribution scenarios, UAC outperforms all baseline methods across every dataset (see \cref{tab:cnn-mix-wisdm-baselines-in}, \cref{tab:cnn-mix-samosa-baselines-in}, and \cref{tab:cnn-mix-uschad-baselines-in}).
Notably, on the Widsm dataset, UAC achieves a $9\%$ improvement in accuracy over the second-best method (Bayesian NN).
For the Samosa and Uschad datasets, the accuracy improvements are $20\%$ and $6\%$, respectively, compared to the second-best method.
Additionally, our method also improves the calibration metrics on all datasets, showing a significant improvement over baseline methods.

In summary, UAC shows improved accuracy and calibration compared to the baseline methods in both in-distribution and out-of-distribution scenarios.

\begin{table}[t]
    \centering
    \caption{Comparison of UAC (ours) with $\text{UAC}_{\neg \sigma}$  on Wisdm, Uschad, and Samosa datasets in out-of-distribution scenario.}
    \label{tab:cnn-sw-weighted-avg-unc-out}
    \resizebox{\columnwidth}{!}{
        \begin{tabular}{lcccccc}
            \toprule
            \multirow{1}{*}{Dataset} & Method                     & Accuracy$\uparrow$       & ECE$\downarrow$            & NLL$\downarrow$            \\
            \midrule
            \multirow{2}{*}{Wisdm}   & $\text{UAC}_{\neg \sigma}$ & $0.64 \pm 0.06$          & $0.138 \pm 0.067$          & $1.280 \pm 0.315$          \\
                                     & UAC (ours)                 & $\mathbf{0.76 \pm 0.08}$ & $\mathbf{0.099 \pm 0.027}$ & $\mathbf{1.043 \pm 0.557}$ \\ \midrule
            \multirow{2}{*}{Samosa}  & $\text{UAC}_{\neg \sigma}$ & $0.47 \pm 0.04$          & $0.100 \pm 0.045$          & $1.778 \pm 0.156$          \\
                                     & UAC (ours)                 & $\mathbf{0.53 \pm 0.06}$ & $\mathbf{0.072 \pm 0.024}$ & $\mathbf{1.687 \pm 0.269}$ \\ \midrule
            \multirow{2}{*}{Uschad}  & $\text{UAC}_{\neg \sigma}$ & $0.74 \pm 0.02$          & $\mathbf{0.085 \pm 0.034}$ & $\mathbf{0.682 \pm 0.107}$ \\
                                     & UAC (ours)                 & $\mathbf{0.77 \pm 0.03}$ & $0.090 \pm 0.044$          & $0.746 \pm 0.256$          \\
            \bottomrule
        \end{tabular}
    }
\end{table}

\begin{table}[t]
    \centering
    \caption{Comparison of UAC (ours) with $\text{UAC}_{\neg \sigma}$ on Wisdm, Uschad, and Samosa datasets in in-distribution scenario.}
    \label{tab:cnn-sw-weighted-avg-unc-in}
    \resizebox{\columnwidth}{!}{
        \begin{tabular}{lcccccc}
            \toprule
            \multirow{1}{*}{Dataset} & Method                     & Accuracy$\uparrow$        & ECE$\downarrow$            & NLL$\downarrow$            \\
            \midrule
            \multirow{2}{*}{Wisdm}   & $\text{UAC}_{\neg \sigma}$ & 0.80 $\pm$ 0.03           & 0.082 $\pm$ 0.019          & 0.675 $\pm$ 0.077          \\
                                     & UAC (ours)                 & \textbf{ 0.98 $\pm$ 0.01} & \textbf{0.057 $\pm$ 0.008} & \textbf{0.159 $\pm$ 0.023} \\ \midrule
            \multirow{2}{*}{Samosa}  & $\text{UAC}_{\neg \sigma}$ & 0.75 $\pm$ 0.02           & 0.176 $\pm$ 0.015          & 0.998 $\pm$ 0.071          \\
                                     & UAC (ours)                 & \textbf{0.97 $\pm$ 0.01}  & \textbf{0.113 $\pm$ 0.011} & \textbf{0.229 $\pm$ 0.031} \\ \midrule
            \multirow{2}{*}{Uschad}  & $\text{UAC}_{\neg \sigma}$ & 0.89 $\pm$ 0.02           & 0.050 $\pm$ 0.014          & 0.253 $\pm$ 0.040          \\
                                     & UAC (ours)                 & \textbf{0.95 $\pm$ 0.01}  & \textbf{0.040 $\pm$ 0.002} & \textbf{0.120 $\pm$ 0.006} \\
            \bottomrule
        \end{tabular}
    }
\end{table}

\begin{table}[t]
    \centering
    \caption{Comparison of the sample prediction using $M_u$, averaging and entropy weighted expectation for all datasets in out-of-distribution scenario.
        % Note that $M_u$ (no $\sigma$) has the same encoder + $M_c$ architecture as $M_u$, however there is not uncertainty estimation, the logits are passed directly through Softmax function.
    }
    \label{tab:entropy-out}
    \resizebox{\columnwidth}{!}{%
        \begin{tabular}{lcccc}
            \toprule
            Dataset                 & Method     & Accuracy$\uparrow$       & ECE$\downarrow$               & NLL$\downarrow$               \\ \midrule

            % encoder$+M_c$ (no $\sigma$) & $0.54 \pm 0.06$ & $0.076 \pm 0.068$ & $1.439 \pm 0.212$ \\
            \multirow{3}{*}{Wisdm}  & $M_u$      & $0.66 \pm 0.06$          & $\mathbf{0.095 \pm 0.058}$    & $1.331 \pm 0.327$             \\
            % UAC (no $\sigma$, no ent) & $0.64 \pm 0.06$ & $0.166 \pm 0.076$ & $1.351 \pm 0.308$ \\
                                    & $M_{avg}$  & $\mathbf{0.75 \pm 0.09}$ & $0.123 \pm 0.035$             & \underline{$1.118 \pm 0.563$} \\
            % UAC (no $\sigma$) & $0.63 \pm 0.06$ & $0.140 \pm 0.071$ & $1.321 \pm 0.311$ \\
                                    & UAC (ours) & $\mathbf{0.75 \pm 0.09}$ & \underline{$0.103 \pm 0.027$} & $\mathbf{1.098 \pm 0.569}$    \\

            \midrule

            % $M_u$ (no $\sigma$) & $0.44 \pm 0.05$ & $0.075 \pm 0.040$ & $1.883 \pm 0.184$ \\
            \multirow{3}{*}{Samosa} & $M_u$      & $0.46 \pm 0.05$          & $0.128 \pm 0.070$             & $2.023 \pm 0.409$             \\
            % UAC (no $\sigma$, no ent) & $0.47 \pm 0.04$ & $0.104 \pm 0.052$ & $1.803 \pm 0.151$ \\
                                    & $M_{avg}$  & $\mathbf{0.53 \pm 0.06}$ & \underline{$0.074 \pm 0.025$} & \underline{$1.697 \pm 0.261$} \\
            % UAC (no $\sigma$) & $0.47 \pm 0.04$ & $0.100 \pm 0.045$ & $1.778 \pm 0.156$ \\
                                    & UAC (ours) & $\mathbf{0.53 \pm 0.06}$ & $\mathbf{0.072 \pm 0.024}$    & $\mathbf{1.687 \pm 0.269}$    \\
            \midrule

            % $M_u$ (no $\sigma$) & $0.70 \pm 0.02$ & $0.074 \pm 0.039$ & $0.826 \pm 0.126$ \\
            \multirow{3}{*}{Uschad} & $M_u$      & $0.72 \pm 0.04$          & $\mathbf{0.087 \pm 0.061}$    & $1.027 \pm 0.383$             \\
            % UAC (no $\sigma$, no ent) & $0.74 \pm 0.02$ & $0.091 \pm 0.025$ & $0.691 \pm 0.110$ \\
                                    & $M_{avg}$  & $\mathbf{0.77 \pm 0.03}$ & $0.097 \pm 0.044$             & \underline{$0.753 \pm 0.256$} \\
            % UAC (no $\sigma$) & $0.74 \pm 0.02$ & $0.085 \pm 0.034$ & $0.682 \pm 0.107$ \\
                                    & UAC (ours) & $\mathbf{0.77 \pm 0.03}$ & \underline{$0.090 \pm 0.044$} & $\mathbf{0.746 \pm 0.256}$    \\

            \bottomrule
        \end{tabular}
    }
\end{table}

\begin{table}[t]
    \centering
    \caption{Comparison of the sample prediction using $M_u$, averaging and entropy weighted expectation for all datasets in in-distribution scenario.}
    \label{tab:entropy-in}
    \resizebox{\columnwidth}{!}{%
        \begin{tabular}{lcccc}
            \toprule
            Method                  & Accuracy$\uparrow$ & ECE$\downarrow$           & NLL$\downarrow$                                               \\ \midrule

            % encoder$+M_c$ (no $\sigma$) & $0.77 \pm 0.02$ & $0.032 \pm 0.005$ & $0.730 \pm 0.086$ \\
            \multirow{3}{*}{Widsm}  & $M_u$              & $0.96 \pm 0.001$          & $\mathbf{0.017 \pm 0.002}$    & $\mathbf{0.126 \pm 0.018}$    \\
            % UAC (no $\sigma$, no ent) & $0.80 \pm 0.03$ & $0.100 \pm 0.023$ & $0.704 \pm 0.076$ \\
                                    & $M_{avg}$          & $\mathbf{0.98 \pm 0.001}$ & $0.068 \pm 0.001$             & $0.172 \pm 0.024$             \\
            % UAC (no $\sigma$) & $0.80 \pm 0.03$ & $0.082 \pm 0.019$ & $0.675 \pm 0.077$ \\
                                    & UAC (ours)         & $\mathbf{0.98 \pm 0.001}$ & \underline{$0.057 \pm 0.008$} & \underline{$0.159 \pm 0.023$} \\

            \midrule

            % $M_u$ (no $\sigma$) & $0.70 \pm 0.02$ & $0.106 \pm 0.017$ & $1.023 \pm 0.061$ \\
            \multirow{3}{*}{Samosa} & $M_u$              & $0.93 \pm 0.01$           & $\mathbf{0.049 \pm 0.004}$    & $0.250 \pm 0.031$             \\
            % UAC (no $\sigma$, no ent) & $0.74 \pm 0.02$ & $0.198 \pm 0.015$ & $1.038 \pm 0.070$ \\
                                    & $M_{avg}$          & $\mathbf{0.97 \pm 0.005}$ & $0.129 \pm 0.012$             & \underline{$0.248 \pm 0.032$} \\
            % UAC (no $\sigma$) & $0.75 \pm 0.02$ & $0.176 \pm 0.015$ & $0.998 \pm 0.071$ \\
                                    & UAC (ours)         & $\mathbf{0.97 \pm 0.005}$ & \underline{$0.113 \pm 0.011$} & $\mathbf{0.229 \pm 0.031}$    \\

            \midrule
            % $M_u$ (no $\sigma$) & $0.90 \pm 0.02$ & $0.022 \pm 0.005$ & $0.220 \pm 0.005$ \\
            \multirow{3}{*}{Uschad} & $M_u$              & $\mathbf{0.95 \pm 0.003}$ & $\mathbf{0.009 \pm 0.006}$    & $\mathbf{0.100 \pm 0.006}$    \\
            % UAC (no $\sigma$, no ent) & $0.90 \pm 0.02$ & $0.057 \pm 0.010$ & $0.250 \pm 0.049$ \\
                                    & $M_{avg}$          & $\mathbf{0.95 \pm 0.005}$ & $0.042 \pm 0.002$             & $0.123 \pm 0.006$             \\
            % UAC (no $\sigma$) & $0.90 \pm 0.02$ & $0.049 \pm 0.008$ & $0.240 \pm 0.048$ \\
                                    & UAC (ours)         & $\mathbf{0.95 \pm 0.005}$ & \underline{$0.040 \pm 0.002$} & \underline{$0.120 \pm 0.006$} \\
            \bottomrule
        \end{tabular}
    }
\end{table}

\subsection{Ablation Studies}

\subsubsection{Uncertainty-aware classification}

To demonstrate the efficiency of the uncertainty prediction, we conduct an ablation study where we compare the results of UAC against using a simple classifier network (encoder + $M_c$)---thus trained without uncertainty estimation---followed by the entropy-weighted expectation.
In the remainder of this paper, we refer to this network as $\text{UAC}_{\neg \sigma}$.

\cref{tab:cnn-sw-weighted-avg-unc-out} and \cref{tab:cnn-sw-weighted-avg-unc-in} summarize the performance across the Wisdm, Uschad, and Samosa datasets, in out-of-distribution and in-distribution scenarios.
% As shown in \cref{tab:cnn-sw-weighted-avg-unc}, modeling uncertainty with UAC consistently yields higher accuracy compared to the baseline.
% Among the datasets, Wisdm contains a large number of test subjects and presents the greatest challenge because each gesture is performed once by each subject, unlike other datasets where subjects perform multiple repetitions of each gesture
In the OOD scenario, the Wisdm dataset shows the most significant improvement among all datasets; UAC achieves an accuracy of $0.76$, compared to $0.64$ with $\text{UAC}_{\neg \sigma}$, marking a $12\%$ improvement.
The Samosa dataset shows a $6\%$ increase, while the USCHAD dataset shows a $3\%$ increase.
In the in-distribution scenario, UAC improves accuracy by $18\%$ on the Widsm dataset, by $22\%$ for the Samosa dataset, and $6\%$ for the USCHAD dataset.

Looking at the calibration metrics, in the OOD scenario, on the Widsm dataset, UAC achieves a decrease in ECE and NLL of respectively $28\%$ and $19\%$.
On the Samosa dataset, the decrease in ECE and NLL are of $22\%$ and $5\%$.
The sole exception is the USCHAD dataset, where incorporating uncertainty results in marginally lower calibration metrics.
However, this difference in calibration is not significant.
In in-distribution scenario, UAC consistently improves both the ECE and NLL---respectively $30\%$ and $76\%$ for the Widsm dataset, $36\%$ and $77\%$ for the Samosa dataset, and $20\%$ and $53\%$ for the USCHAD dataset.
% Similarly, for the Samosa dataset, UAC achieves better calibration metrics than UAC (no $\sigma$)---with an ECE of $0.072$ and a NLL of $1.687$.
% However, for the Uschad dataset, UAC achieve almost comparable ECE to UAC (no $\sigma$) with $0,090$ and $0.085$ respectively.
% A similar conclusion can be drawn for the NLL---0.682 and 0.746 respectively.
% The ECE and NLL over the three test dataset show that UAC improves or maintains calibration while improving prediction accuracy.

% On all datasets and scenarios, UAC consistently outperforms $\text{UAC}_{\neg \sigma}$ highlighting the efficacy of integrating predictions' uncertainties for improved accuracy and calibration.

% The comparison with the identical architecture without uncertainty prediction---UAC~(no~$\sigma$)---underscores the significance of incorporating uncertainty prediction into the prediction strategy.

On all datasets and scenarios, UAC consistently outperforms $\text{UAC}_{\neg \sigma}$.
The comparison with the identical architecture without uncertainty prediction---$\text{UAC}_{\neg \sigma}$---underscores the significance of incorporating uncertainty prediction into the prediction strategy for improved accuracy and calibration.

\subsubsection{Entropy-weighted expectation}

To demonstrate the effectiveness of the entropy-weighted expectation in improving accuracy while preserving the calibration of the model, we perform an ablation study.
This study compares the performance of UAC with $M_u$ (as described in \cref{fig:method:step1}) and a variant of UAC, referred to as $M_{avg}$, where the entropy-weighted expectation is replaced with a simple arithmetic mean.

As seen in \cref{tab:entropy-out}, UAC maintains the performance improvements in accuracy achieved through $M_{avg}$'s averaging---$9\%$ on the Widsm dataset, $7\%$ on the Samosa dataset, and $5\%$ on the Uschad dataset---while improving model calibration.
The expectation over multiple samples reduces the influence of random noise and the impact of sample misclassifications, leading to improved performance.
On the other hand, when comparing UAC with $M_u$, the ECE increases slightly while the NLL decreases, except for the Samosa data set.
The increase in ECE suggests a tendency of the model to assign slightly higher probabilities to false positives.
However, the simultaneous decrease in NLL suggests improved accuracy and confidence in correct predictions.
Given that the NLL improves and that the ECE increase is minimal, using UAC for its improved accuracy and calibration is advantageous in OOD scenarios.

Looking at the in-distribution scenario in \cref{tab:entropy-in}, the results are less conclusive.
While, as in OOD scenario, averaging methods (UAC or $M_{avg}$) improve accuracy and UAC improves the calibration compared to $M_{avg}$, $M_u$ generally has better calibration.
Thus, in the in-distribution scenario, there is a trade-off between improving accuracy and maintaining model calibration.
The choice between UAC and the uncertainty-weighted single sample prediction network $M_u$ should be based on the priority given to either accuracy or calibration.

\subsection{Computational Efficiency}

Our model was trained and tested using a single Nvidia T4 GPU.
For the Wisdm dataset---the largest dataset in our experiments---training the first step of UAC required 10 hours and 42 minutes.
However, inference is step 2 was nearly instantaneously.
Step 2, involving calculating the expectation of multiple IMU measurements, was computed in just 0.001 seconds for 25 samples.
This demonstrates the computational efficiency of our approach.
% , making it suitable for real-time applications.

\section{Conclusion and Future Work}
\label{sec:conclusion}

The adoption of machine learning in safety-critical environments, such as construction sites, remains limited due to the need for guaranteed system safety and reliability.
Additionally, privacy concerns often restrict the use of certain sensors, favoring Inertial Measurement Units (IMUs) over cameras.
Therefore, gesture detection algorithms for safety-critical applications must be both accurate and well-calibrated, even when relying solely on IMU data.

In this paper, we introduce a method for gesture detection using IMU data, focusing on enhancing both prediction accuracy and model calibration.
Our approach, named UAC, operates in two stages.
First, a neural network is trained to, from sample motion sequence data, predict both the probabilities associated with each possible label and the uncertainty of the prediction.
Second, using the predicted uncertainty, the initial probabilities are calibrated, and accuracy is further improved by performing the entropy-weighted expectation over multiple samples extracted from a gesture sequence.

Our experiments, across three datasets and against three state-of-the-art uncertainty and calibration baselines (entropy-maximization, temperature scaling, and Bayesian neural networks), demonstrate that our method achieves improved accuracy and calibration in both in-distribution and out-of-distribution scenarios.
Furthermore, ablation studies highlight the critical role of uncertainty prediction and entropy-weighted expectation in our approach.
However, while UAC is generally better than the state-of-the-art in OOD scenarios, we show that, in in-distribution scenarios, there exists a trade-off between accuracy and calibration when using entropy-weighted expectation in our approach.

A limitation of our approach lies in the sampling strategy used to extract samples from a motion sequence of IMU data.
The use of a sliding window may not consistently capture samples that are representative of the executed motion.
Future research will aim to develop more efficient sampling techniques.
Furthermore, while our focus is on IMU-based gesture recognition, future work will focus on exploring the generalization capabilities of UAC to other sensor modalities---e.g. piezoresistive sensors that measure muscle contraction.
Lastly, althought we demonstrated real-time capabilities, it still requires the use of GPU.
Therefore, future work should focus on developing a lightweight version of UAC that can be deployed on edge devices.

% \section*{Acknowledgments}
% This research was supported by Schindler AG.

\printbibliography
\end{document}